\documentclass[letterpaper, 10 pt, conference]{ieeeconf}
\usepackage{amsmath,amsfonts}
\usepackage{array}
\usepackage{enumerate}
\usepackage[utf8]{inputenc}
\usepackage[T1]{fontenc}
\usepackage{amsfonts}
\usepackage{amssymb}
\IEEEoverridecommandlockouts 
\usepackage{algorithm}  
\usepackage{algpseudocode}  
\usepackage{amsmath}  
\usepackage{breqn}
\usepackage{textcomp}
\usepackage{stfloats}
\usepackage{url}
\usepackage{verbatim}
\usepackage{graphicx}
\usepackage{cite}
\usepackage{color}
\usepackage{colortbl}
\usepackage{hyperref}
\hypersetup{
    colorlinks=true,
    linkcolor=black,
    filecolor=black,      
    urlcolor=black,
    }

\title{\LARGE \bf Initial Task Allocation for Multi-Human Multi-Robot Teams \\ with Attention-based Deep Reinforcement Learning  

}

\author{Ruiqi Wang$^{1}$, Dezhong Zhao$^{1,2}$, and Byung-Cheol Min$^{1}$ 
\thanks{$^{1}$SMART Laboratory, Department of Computer and Information Technology, Purdue University, West Lafayette, IN, USA. {\tt\small{[wang5357,minb]@purdue.edu}.}}
\thanks{$^{2}$College of Mechanical and Electrical Engineering, Beijing University of Chemical Technology, Beijing, China. \tt\small{DZ\_Zhao@buct.edu.cn}.}}

\begin{document}

\maketitle

\begin{abstract}
Multi-human multi-robot teams have great potential for complex and large-scale tasks through the collaboration of humans and robots with diverse capabilities and expertise. To efficiently operate such highly heterogeneous teams and maximize team performance timely, sophisticated initial task allocation strategies that consider individual differences across team members and tasks are required. While existing works have shown promising results in reallocating tasks based on agent state and performance, the neglect of the inherent heterogeneity of the team hinders their effectiveness in realistic scenarios. In this paper, we present a novel formulation of the initial task allocation problem in multi-human multi-robot teams as a contextual multi-attribute decision-make process and propose an attention-based deep reinforcement learning approach. We introduce a cross-attribute attention module to encode the latent and complex dependencies of multiple attributes in the state representation. We conduct a case study in a massive threat surveillance scenario and demonstrate the strengths of our model.
\end{abstract}

%Such human-robot teams have realized tremendous success in various domains due to the leverage of the strengths of both humans, e.g., situational awareness, and robots, e.g., consistent work performance.

\section{Introduction}
Human-robot teams offer significant benefits across various domains by combining the strengths of autonomous systems such as consistency and preciseness, with the creativity and adaptability associated with humans \cite{ajoudani2018progress}. The increasing demand for more efficient solutions to complex and large-scale tasks, such as disaster response, search and rescue, and environmental surveillance, has spurred the interest in multi-human multi-robot (MH-MR) teams \cite{dahiya2023survey}. The collaboration of multiple humans and robots with diverse capabilities, expertise, and characteristics involved in an MH-MR team presents great potential to enhance team complementarity, productivity, and versatility \cite{ramchurn2015study,freedy2008multiagent}. Nevertheless, the high heterogeneity of such a team also leads to increased operational challenges in efficiently coordinating agents. Therefore, optimal task allocation across heterogeneous agents is a crucial challenge that must be addressed to fully realize the potential of an MH-MR team.

%H\cite{ajoudani2018progress,esterwood2020human}. Previously, most research has focused on dyadic human-robot teams, where a single human interacts with a single robot. However, the increasing demand for more efficient solutions to complex and large-scale tasks, such as disaster response, search and rescue, and environmental surveillance, has led to a shift towards multi-human multi-robot (MH-HR) teams \cite{dahiya2023survey}. 

While task allocation has been well-researched in the context of multi-robot systems \cite{khamis2015multi,gini2017multi,creech2021resource} and human-autonomy collaboration \cite{karami2020task,johannsmeier2016hierarchical,wu2022task},  scenarios involving MH-MR teams have received relatively little attention. In the limited literature available, most works only focus on in-process task distribution, which utilizes human and robot states, as well as task performance metrics, as indicators. However, these studies often overlook the initial task scheduling that considers individual differences of humans, robots, and tasks \cite{mina2020adaptive,yu2021optimizing,ham2021human,patel2020improving}. This initial task arrangement significantly impacts team performance. Correct initial allocation can leverage the intrinsic heterogeneity of an MH-MR team by ensuring that individuals are assigned tasks that match their abilities and skillset at the start, leading to maximum performance in a shorter period of time. Conversely, incorrect settings hinder the team's ability to optimize performance through in-process task reallocation, even after a prolonged period of time.

\begin{figure}[t]
\centering
\includegraphics[width=0.95\columnwidth]{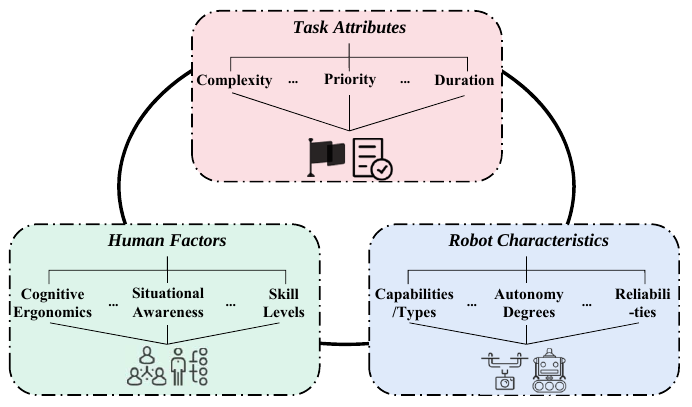}
\vspace{-2pt}
\caption{Illustration of the contextual multi-attribute decision-making process used to formulate the initial task allocation problem in an MH-MR team.}
\label{fig:concept}
\vspace{-15pt}
\end{figure}

To fill the gap in the literature, we aim to explore the problem of initial task allocation (ITA) in MH-MR teams. Specifically, we seek to determine how to optimally assign a job consisting of a set of tasks, each with varying attributes, to a team of humans with diverse capabilities and robots with assorted characteristics at the outset. As illustrated in Fig. \ref{fig:concept}, we approach this problem by formulating it as a contextual multi-attribute decision-making process (CMADP) \cite{xu2015uncertain}. This involves making a decision by considering a context consisting of three main categories of attributes: human factors, robot characteristics, and task attributes, each of which includes various sub-attributes composed of individual variables \cite{tabrez2020survey}. However, finding the optimal decision for ITA is challenging for two main reasons. Firstly, the decision is not solely driven by diverse attributes and sub-attributes, but rather by the latent dependencies and interactions across them \cite{ma2022metrics}. Secondly, the importance of different relations is dynamic, varying under different job scenarios.

\begin{figure*}[!t]
\centering
\includegraphics[width=1\linewidth]{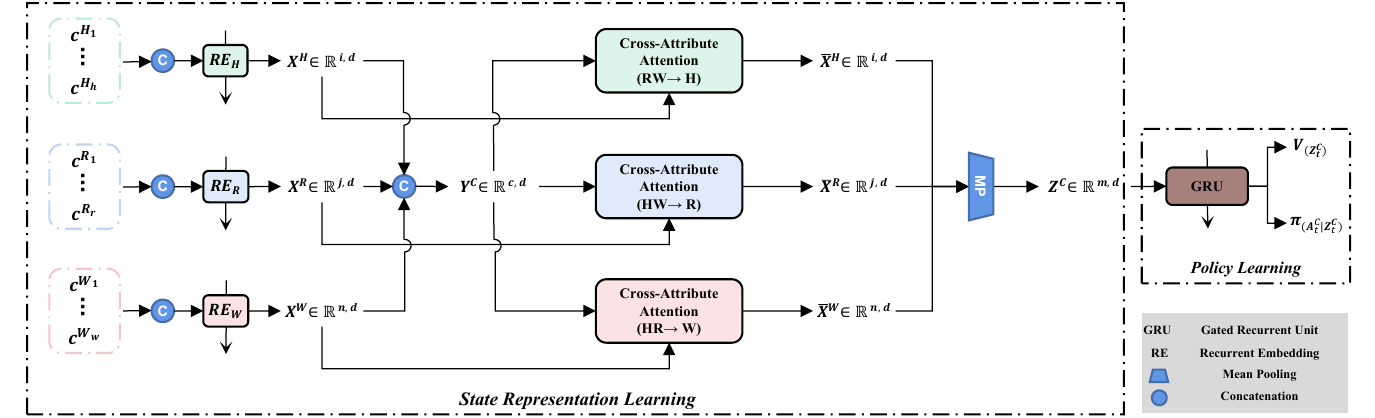}

\caption{Illustration of the proposed AtRL framework. The multi-attribute data inputs consisting of $c^{(.)_{(*)}} \in \mathbb{R}^{{L},{D}}$, where $c^{(.)_{(*)}}$ presents the $*^{th}$ sub-attribute sequence within the $.^{th}$ attribute category, and $L$ and $D$ respectively denote the length and dimension of the sequence, are firstly fed into recurrent embedding layers $RE_{(.)}$ separately to generate three attribute sequences $X^{(.)} \in \mathbb{R}^{{L_{(.)}},{d}}$ with the same dimension $d$ (Section \ref{RE}). Such sequences are concatenated to generate a low-level state representation $Y^{C} \in \mathbb{R}^{{{c}},{d}}$, which is passed through cross-attribute attention layers with each attribute sequence $X^{(.)}$ respectively. In each cross-attribute attention layer, each attribute sequence is adapted with relative information revealed in the other two, by calculating adaptive dependencies between features of the current attribute and those encoded in the $Y^{C}$ (Section \ref{SR}). Finally, the enhanced attribute sequences $\overline{X}^{(.)}$ are then passed through a mean pooling layer to produce the high-level state presentation of the multi-attribute context $Z^{C}\in \mathbb{R}^{{m},{d}}$, which is transported to a policy network to learn the value function $V$and policy $\pi$ (Section \ref{PL}).}
\vspace{-10pt}
\label{fig:framwork}
\end{figure*}

To address these challenges, we propose an \textbf{At}tention-based Deep \textbf{R}einforcement \textbf{L}earning approach (AtRL). The architecture and procedure of the AtRL are illustrated in Fig. \ref{fig:framwork}. At the core of our model is cross-attribute attention state representation learning, which aims to learn a global state representation of the multi-attribute context with adaptive correlation information encoded. Specifically, cross-attribute attention layers adaptively capture the latent dependencies across different attribute categories, continually enhancing features within one certain attribute with strongly relevant information from the other two. Then, the enhanced attribute feature sequences pass through a mean pooling layer to produce the global state representation. To evaluate the performance of our AtRL and the cross-attribute attention inside, we conduct a case study in a large-scale threat surveillance and identification task scenario.
%Each column of each attribute sequence presents a sum of convoluted sub-attributes of each human agent, robot, or task. concatenation
 
%Each attribute sequence has different lengths, e.g., the number of human agents, but the same dimension, e.g., the number   

%Namely, features of each main attribute are mapped as an attribute sequence with different lengths, e.g., the number of human agents, and dimensions, e.g., the number   

The main contributions can be summarized as followings: 1) We conduct pioneering research to investigate the initial task allocation problem in MH-MR teams and provide a problem formulation for the contextual multi-attribute decision-making process; 2) To tackle challenges associated with the CMADP, we propose a novel learning-based framework with a cross-attribute attention representation learning module to adaptively capture the latent dependencies in the multi-attribute context; and 3) Our case study, which involved a large-scale threat surveillance task scenario, demonstrates the effectiveness of our AtRL and the cross-attribute attention inside.

\section{Background and Preliminary}

\noindent \textbf{Related Works.} A multi-human, multi-robot (MH-MR) team is a group consisting of multiple human agents and multiple robots working together to achieve a shared objective comprised of different tasks \cite{dahiya2023survey}. Such a team is highly heterogeneous in nature for the significant differences that exist among the human agents and the robots in terms of their capabilities, skills, and characteristics. For instance, human agents obtain varied levels of cognitive resistance to fatigue and workload, and different operational skills and decision-making abilities. Also, the robots in an MH-MR team often have varying types, such as the combination of unmanned aerial vehicles (UAVs) and unmanned ground vehicles (UGVs) in field search and rescue operations, resulting in different robotic characteristics. To take advantage of such high heterogeneity and maximize team performance, sophisticated initial task allocation strategies are required to ensure that individuals are assigned tasks that match their abilities and skill sets from the outset.

Unfortunately, in the limited literature, most existing workload distribution methods for MH-MR teams neglect this intrinsic heterogeneity, only considering in-process workload reallocation based on the states and performance of each human agent and robot \cite{mina2020adaptive,yu2021optimizing,ham2021human,patel2020improving}. For example, Mina \textit{et. al} \cite{mina2020adaptive} developed an adaptive workload allocation system that monitors human cognitive workload, robot conditions, and individual performance to adjust the workload accordingly, ensuring that each agent is in an optimal work state. Despite the promising results of such in-process task allocation methods, failing to consider the fundamental heterogeneity in MH-MR teams can reduce their effectiveness in realistic scenarios. In particular, if the initial workload distribution is incorrect, optimizing team performance through in-process task reallocation becomes difficult, even after a prolonged period of time. On the other hand, few studies consider the task distribution for MH-MR teams at the beginning. For instance, Humann \textit{et. al} \cite{humann2018modeling} builds a model of a multi-UAV, multi-operator surveillance system, simulating the impact of the number of human operators and two types of UAVs on team performance. However, it fails to model individual differences in humans adequately and, more importantly, does not provide any methods to optimize the initial workload distribution. To address these gaps, we explore the optimal initial task allocation that takes into account the inherent heterogeneity of the MH-MR team. We provide a problem formulation of a contextual multi-attribute decision-making process and propose an attention-based reinforcement learning solution to this problem.

\noindent \textbf{State Representation Learning.} 
State representation learning is a critical problem in reinforcement learning that involves automatically extracting useful and compact representations of the environment states to facilitate the policy learning process, especially when the state space has a large and complex structure \cite{de2018integrating}. Recent research has shown that attention-based neural networks, such as graph attention networks and Transformer layers \cite{ni2021multi}, are effective in reasoning the state space by mapping raw sensory inputs to a low-dimensional state representation that captures the most relevant features of the environment. The learned state representation can then be used as input to a policy network to learn an optimal policy.

\noindent \textbf{Proximal Policy Optimization.} In this work, Proximal Policy Optimization (PPO)\cite{schulman2017proximal} is utilized as the training algorithm. The objective of policy optimization in PPO can be formulated as a weighted sum of three loss terms: the clipped surrogate objective for policy improvement ($\mathcal{L}^{CP}$), a value function loss ($\mathcal{L}^{V}$), and an entropy loss ($\mathcal{L}^{O}$) for regularization, formally as:
\begin{equation}
\label{PPO}
\mathcal{L}(\theta, \phi) =w_p \mathcal{L}^{CP}(\theta)- w_v \mathcal{L}^{V}(\phi)+ w_e \mathcal{L}^{O}(\theta)    
\end{equation}
\noindent with
\begin{equation}
\label{PPO2}
\begin{aligned}
\mathcal{L}^{CP}(\theta)&=\mathbb{E}_t[\sum_0^t \min \big(r_{t}(\theta) \hat{A}_{t}, \\ &\operatorname{clip}\left(r_{t}(\theta), 1-\epsilon, 1+\epsilon\right) \hat{A}_{t}\big)] \\ 
\mathcal{L}^{V}(\phi)&=\mathbb{E}_t[\sum_0^t\left(V_\phi\left(s_{t}\right)-\hat{A}_{t}\right)^2]  \\
\mathcal{L}^{O}(\theta)&=\mathbb{E}_t[\sum_0^t \mathcal{O}\left(\pi_\theta\left(a_{t} \mid s_{t}\right)\right)]  \\
\end{aligned}
\end{equation}

\noindent where $w_p$, $w_v$, and $w_e$ denote the loss coefficients for the clipped policy loss, value function loss, and entropy loss respectively. $\hat{A}$ represents estimates of the advantage function, $\epsilon$ is the clipping ratio, $\mathcal{O}(\cdot)$ refers to the entropy, and $r_t(\theta)$ is the importance sampling ratio as:
\begin{equation}
\label{PPO3}
r_t(\theta) = \frac{\pi_\theta(a_t|s_t)}{\pi_{\theta_{\text{old}}}(a_t|s_t)}  
\end{equation}
\noindent where $\pi_{\theta_{\text{old}}}$ is the previous policy.

\section{Problem Formulation}
\label{PF}
We formulate the initial task allocation (ITA) problem in an MH-MR team as a contextual multi-attribute decision-making process (CMADP), which is defined as a tuple $({W}, \overline{\alpha}, {C}, {A}, \mathcal{T}, {R})$, where:

\begin{itemize}
    \item ${W}$ := \{${w_1}$, \dots, ${w_n}$\} is a finite set of $n$ tasks with varying attributes
    \item $\overline{\alpha}$ := \{${\alpha_1}$, \dots, ${\alpha_i}$, ${\overline{\alpha}_1}$, \dots, ${\overline{\alpha}_j}$\} is a finite set of $i$ human agents with diverse capabilities and $j$ robots with assorted characteristics
    \item ${C}$ := \{${C^{H}}$ $\times$ ${C^{R}}$ $\times$ ${C^{W}}$\} is the joint multi-attribute context observed, including joint factors of $i$ human agents measured in $h$ dimensions, ${C^{H}}$ := \{\{${c^{H_1}_1}$, \dots, ${c^{H_1}_i}$\} $\times$ \dots $\times$ \{${c^{H_h}_1}$, \dots, ${c^{H_h}_i}$\}\}, joint characteristics of $j$ robots presented in $r$ dimensions, ${C^{R}}$ := \{\{${c^{R_1}_1}$, \dots, ${c^{R_1}_j}$\} $\times$ \dots $\times$ \{${c^{R_r}_1}$, \dots, ${c^{R_r}_j}$\}\}, and joint attributes of $n$ tasks assessed in $w$ dimensions, ${C^{W}}$ := \{\{${c^{W_1}_1}$, \dots, ${c^{W_1}_n}$\} $\times$ \dots $\times$ \{${c^{W_w}_1}$, \dots, ${c^{W_w}_n}$\}\}
    \item ${A}$ := \{${a_1}$ $\times$ \dots $\times$ ${a_n}$\} is the joint allocation decision action by assigning $w_n$ to $\alpha_i$ and/or ${\overline{\alpha}_j}$
    \item $\mathcal{T}$ := ${Pr}\left({C}^{\prime} \mid {C} \right)$ is a random contextual observation transition probability
    \item ${R}$ := $f_R\left({C},{A}\right)$ = $\mathbb{E}[R_t(C_t, A_t)]$ = $\mathbb{E}[\sum_{k=1}^n r_k(c_t, a_t)]$ is the reward function that gives an immediate reward $R_t$ after taking a joint action $A_t$ when observing a joint context $C_t$ at time step $t$. It is the sum of the individual reward $r_n$ for each assigned task $n$, reflecting the individual task performance of task $n$. 
\end{itemize}

Similar to the contextual multi-armed bandit problem \cite{zhang2022feel}, the CMADP can be viewed as a repeated game between a player (ITA policy) and an adversary (context selector). At each time step $t$, the adversary selects a multi-attribute context ${C_t}\in{C}$ based on the game history $H_{t-1}=\left[\left(C_1, A_1, R_1\right), \ldots,\left(C_{t-1}, A_{t}, R_{t-1}\right)\right]$, and the same contexts can be selected multiple times. The player selects an ITA action ${A_t}\in{A}$ based on its policy $\pi (A_t|C_t)$ and receives an immediate reward $R_t$, which only depends on (${C_t},{A_t}$). The goal of the player is to find the optimal policy $\pi^{*}: \mathbf{C}_{t} \mapsto \mathbf{A}_{t}$ that maximizes the expected reward $\mathbb{E} [\sum_{s=1}^t f_R\left(C_s, A_s\right)]$. 

This formulation enables the fundamental modeling of contextual dynamics of the ITA problem in an MH-MR team, and explicitly defines various attributes that influence the decision-making process, including human factors, robotic characteristics, and task-specific factors. It is worth noting that our CMADP differs from the traditional Markov decision process (MDP) in that  the initial workload decision action does not impact the context (state), which presents the inherent heterogeneity of the team and tasks. However, the contextual observation transition probability $\mathcal{T}$, introduced by the notion of the adversary, serves as a similar function to the state transition function.

\section{Methodology}
In this section, we introduce our proposed AtRL for initial task allocation in MH-MR teams.
\subsection{Overview}
 Our goal is to solve the contextual multi-attribute decision-making problem formulated in the previous section and train an allocation policy that can determine the optimal initial task distribution based on the inherent heterogeneity of human agents, robots, and tasks, which is defined as a multi-attribute context. However, since the state space of a multi-attribute context is highly complex and implicitly includes dependencies across different attributes and sub-attributes, it is crucial to learn a state representation that can efficiently and adaptively model such latent interactions. To this end, we introduce cross-attribute attention for state representation learning, which can provide a better understanding of the multi-attribute context and accelerate the policy learning process. 

\subsection{Recurrent Embedding}
\label{RE}
Consider an MH-MR team consisting of $i$ human agents and $j$ robots assigned to complete $n$ tasks. As described in the problem formulation in Section \ref{PF}, the joint multi-attribute context comprises three main attributes: factors of $i$ human agents measured in $h$ dimensions, characteristics of $j$ robots presented in $r$ dimensions, and attributes of $n$ tasks assessed in $w$ dimensions. 

Formally, we can obtain three raw attribute sequences: $C^{H} \in \mathbb{R}^{{i},{h}}$, $C^{R} \in \mathbb{R}^{{j},{r}}$ and $C^{W} \in \mathbb{R}^{{n},{w}}$ as:
\begin{equation}
\begin{aligned}
C^{H}&=\circ\left(C^{H_1}, C^{H_2}, \dots, C^{H_h}\right) \\
C^{R}&=\circ\left(C^{R_1}, C^{R_2}, \dots, C^{R_r}\right) \\
C^{W}&=\circ\left(C^{W_1}, C^{W_2}, \dots, C^{W_w}\right)
\end{aligned}
\label{eq:4}
\end{equation}

\noindent where $\circ$ denotes the concatenation operation, $C^{H_h} \in \mathbb{R}^{{i},{1}}$, $C^{R_r} \in \mathbb{R}^{{j},{1}}$, and $C^{W_w} \in \mathbb{R}^{{n},{1}}$ refer to the sequence data of sub-attributes within the three attribute classes, respectively.

Then each raw attribute sequence is applied with a nonlinear transformation operation, e.g., Tanh activation function with fully connected layers, and then fed into a RNN cell with $d$ hidden units as:
\begin{equation}
\begin{split}
\hat{C}^{H}_i&=\operatorname{RNN}\left(\hat{C}^{H}_{i-1}, {Tr}\left({C}^{H}_i\right)\right)\\
\hat{C}^{R}_j&=\operatorname{RNN}\left(\hat{C}^{R}_{j-1}, {Tr}\left({C}^{R}_j\right)\right)\\
\hat{C}^{W}_n&=\operatorname{RNN}\left(\hat{C}^{W}_{n-1}, {Tr}\left({C}^{W}_n\right)\right)
\end{split}
\label{eq:5}
\end{equation}

\noindent where $\hat{C}^{H}_i$, $\hat{C}^{R}_j$ and $\hat{C}^{W}_n$ denote the hidden state at the final time step respectively, and $Tr$ is the nonlinear transformation operation. Note that these three RNNs have different input dimensions.

We regard these final hidden states as the low-level attribute sequences $X$ as: $X^H = \hat{C}^{H}_i \in \mathbb{R}^{{i},{d}}$, $X^R = \hat{C}^{R}_j \in \mathbb{R}^{{j},{d}}$, and $X^W = \hat{C}^{W}_n \in \mathbb{R}^{{n},{d}}$. Each of these sequences aims to obtain low-level features of each attribute while taking into account spatial information, which is critical given that the task allocation decision is order-dependent. It is also important to note that after the recurrent embedding, the three attribute sequences share the same dimension $d$, which makes the dot-product operation in the subsequent cross-attribute attention mechanism mathematically feasible.

Then these three attribute sequences are concatenated to produce a low-level multi-attribute state representation $Y^{C} \in \mathbb{R}^{{c},{d}}$ as:
\begin{equation}
Y^{C}=\circ\left(X^{H}, X^{R}, X^{W}\right).
\label{eq:6}
\end{equation}

\subsection{Cross-attribute Attention-enhanced State Representation}
\label{SR}

The purpose of our cross-attribute attention is to comprehensively and adaptively encode the latent correlation information among human factors, robot characteristics, and task attributes into the state representation of the multi-attribute context. Specifically, in a cross-attribute attention layer, an attention score is calculated between an attribute sequence $X$ and the low-level multi-attribute state representation $Y^{C}$, which guides the adaptation of the uni-attribute features by incorporating relevant information from the other two attributes embedded in the state representation. 

In the following, we illustrate how the cross-attribute mechanism works using the attribute sequence of human factors $X^H \in \mathbb{R}^{{i},{d}}$, as the example uni-attribute input. Formally, the query $Q_{H}$, key $K_{C}$ and value $V_{C}$ are defined as: 
\begin{equation}
\label{Cross_qkv}
\begin{aligned}
Q^{H} &= X^{H} \cdot W^{Q^{H}}\\
K^{C} &= Y^{C} \cdot W^{K^{C}}\\
V^{C} &= Y^{C} \cdot W^{V^{C}}
\end{aligned}
\end{equation}
\noindent where $W^{Q^{H}} \in \mathbb{R}^{d, q}$, $W^{K^{C}} \in \mathbb{R}^{d, k}$, and $W^{V^{C}} \in \mathbb{R}^{d, v}$ denote three groups of trainable weights.

\begin{figure}[t]
\centering
\includegraphics[width=0.9\columnwidth]{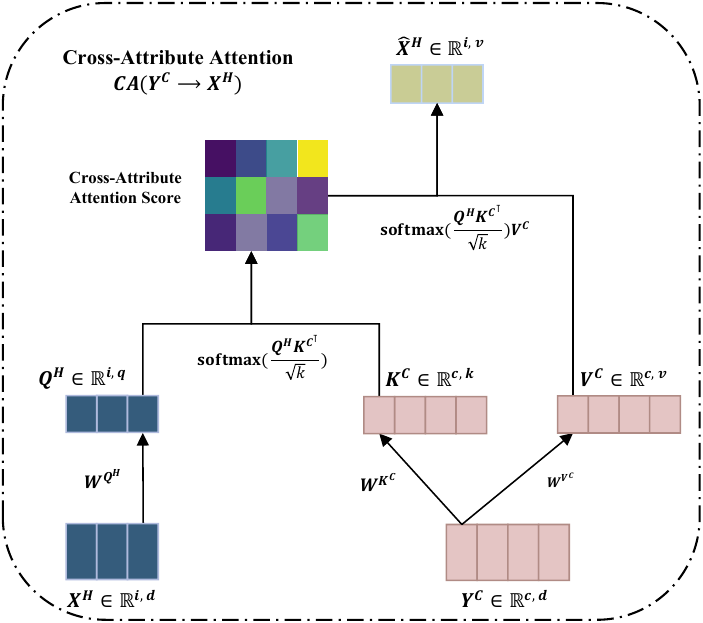}
\vspace{-2pt}
\caption{An example illustration of the cross-attribute attention mechanism.}
\vspace{-15pt}
\label{fig:CA}
\end{figure}

Similar to the self-attention mechanism used in \cite{vaswani2017attention}, our cross-attribute attention allows the flow of relevant information, revealing latent dependencies across attributes, from the multi-attribute representation to the attribute sequence of human factors, as depicted in Fig. \ref{fig:CA}:
\begin{equation}
\label{eq_ca}
\begin{split}
\hat{X}^H &= CA(Y^C \rightarrow X^H)\\
&= {\text{Attention} (Q^{H}, K^{C}, V^{C})}\\
&= \operatorname{softmax}\left(\frac{Q^{H} \cdot {K^C}^\top}{\sqrt{k}}\right){V^C}.
\end{split}
\end{equation}

This process can be performed in parallel multiple times to serve as the multi-head cross-attribute attention. Then the enhanced attribute sequence of human factors $\overline{X}^{H}$ can be obtained as:
\begin{equation}
\overline{X}^{H}=\operatorname{LN}\left(\operatorname{FFN}\left(\hat{X}^H\right)+{X}^{(H)}\right)
\end{equation}
\noindent where LN stands for layer normalization operation and FFN presents feed-forward layers.

Similarly, we can obtain the enhanced feature representations for the other two attribute sequences $\overline{X}^{R}$ and $\overline{X}^{W}$ (robot characteristics and task attributes) by applying the same procedure. Finally, the three sequences are aggregated through a mean pooling layer to generate the high-level state representation of the multi-attribute context $Z^C$ as:
\begin{equation}
f_{\vartheta}\left(Z^C\right)=\operatorname{MP}\left(\overline{X}^{H}, \overline{X}^R, \overline{X}^W\right)
\end{equation}
\noindent where MP stands for the mean pooling operation. For brevity, we denote the learned multi-attribute state representation as $f_{\vartheta}\left(Z^C\right)$, where ${\vartheta}$ represents the set of all trainable parameters in the cross-attribute state representation learning.

This mean pooling layer is helpful as the global state representation can be potentially high-dimensional, especially when dealing with large-sized or highly heterogeneous MH-MR teams and tasks. This layer can map the high-dimensional global state representation concatenated by enhanced attribute sequences to a lower-dimensional space, making it computationally feasible in the following policy learning. 

\subsection{Policy Learning}
\label{PL}

After obtaining the state representation $Z^C$ enhanced by cross-attribute attention, we feed it into the policy network. Due to the sequential and complex nature of the state representation, we utilize gated recurrent units (GRU) instead of traditional fully connected layers as the policy network. This enables the agent to better capture the dynamic patterns in the state representation and learn the optimal ITA policy. 

Following the formulation in Section \ref{PF}, the agent is provided with a learned multi-attribute state representation $Z^C_t$ obtained from cross-attribute learning at each step of the policy learning. Based on the current policy $\pi_\theta$, it selects a joint allocation strategy $A_t$ to assign tasks to each human agent and robot in the team and receives a reward $R_t$ from the reward function $f_R(C, A)$, which evaluates the team's performance under the selected task allocation. The agent optimizes the parameter sets $\theta$, $\phi$ of its policy and value function using PPO to maximize the objective function described in Eq. \ref{PPO}. Additionally, the parameter set $\vartheta$ in the state representation learning is optimized as part of the value function.

%The overall procedure of our AtRL is illustrated in Algorithm \ref{alg:1}.

\iffalse

\begin{algorithm}[h]  
	\caption{AtRL with PPO}  		
	\label{alg:1}
	\begin{algorithmic}[1]  
            \Statex \textbf{Input:} Three raw sequences of $C^H$, $C^R$ and $C^W$
            \Statex \textbf{Initialize:} $c$
		\State Initialize a critic $Q_{\theta}$
		\State // Hybrid Experience Learning
		\State $\pi_{\phi} \leftarrow$ Algorithm \ref{alg:1}
		\State // Agent Learning
		\For{ each time step }
		\State //  Active Reward Learning
		\State $EB, \hat{R}_{\mu}\leftarrow$ Algorithm \ref{alg:2}
		\State Take $a_t \sim \pi_{\phi}(a_t | s_t)$ and reach $s_{t+1}$
		\State Compute reward $\hat{R}_{\mu}(s_t,a_t) \leftarrow \hat{R}_{\mu}$
		\State Store transitions 
		\State ${EB} \leftarrow EB \cup \{(s_t,a_t,s_{t+1},\hat{R}_{\mu}(s_t,a_t))\}$
		\State // Policy Optimization
		\For{each gradient step}
		\State Sample minibatch transitions $\sim EB $ 
		\State Optimize $\theta$, $\phi$ by $\mathcal{L}_Q(\theta)$ in (\ref{eq2}) and $\mathcal{L}_{\pi}(\phi)$ in (\ref{eq4})
		\EndFor
			\EndFor
		\State \Return $\pi_{\phi}$, $Q_{\theta}$
	\end{algorithmic}  
\end{algorithm}  

\fi

\section{A Case Study in Threat Surveillance Scenario}
In this section, we describe our case study to apply and validate our proposed AtRL in a large-scale threat surveillance task scenario\footnote{More details can be found at \small{\url{https://sites.google.com/view/ITA-AtRL}}}.

\subsection{Task Scenario}
The application domain that we consider in our case study is the large-scale surveillance and classification of unknown threats within one area using an MH-MR team. This scenario is common in military and environmental applications, where potential threats, such as enemy equipment or hazardous materials, need to be monitored and identified at a closer distance than satellite imagery.

In our case study, the objective of the MH-MR team is to perform large-scale surveillance and classification of potential threats in a given territory, utilizing a combination of human operators and two types of robots. The task begins with the satellite system detecting and generating a list of several points of interest (POIs). The command system then allocates robots to these reported POIs to gather close-up visual information, while human operators use this information to identify each POI as a potential threat or not. Multiple robots and human operators can work simultaneously to expedite the process.

To represent the heterogeneity of the MH-MR team, we consider multiple attributes of human operators, robots, and tasks. For human factors, we focus on cognitive ability and operational skill level, which are two main factors affecting human performance in a human-robot team \cite{harriott2013modeling}. Cognitive ability measures the resilience of a person to fatigue and workload, while operational skill level refers to an individual's level of skill or expertise in a specific task or set of tasks acquired through training and experience.  Concerning robot characteristics, we consider two types of robots: unmanned aerial vehicles (UAVs) and unmanned ground vehicles (UGVs). UAVs can move quickly and provide a top-down view of POIs from a certain height, while UGVs move more slowly but can provide a clearer view to humans. Finally, we consider the location information of POIs and the difficulty of threat classification as task attributes.

\subsection{Simulation Environment}

\begin{figure}[t]
\centering
\includegraphics[width=0.9\columnwidth]{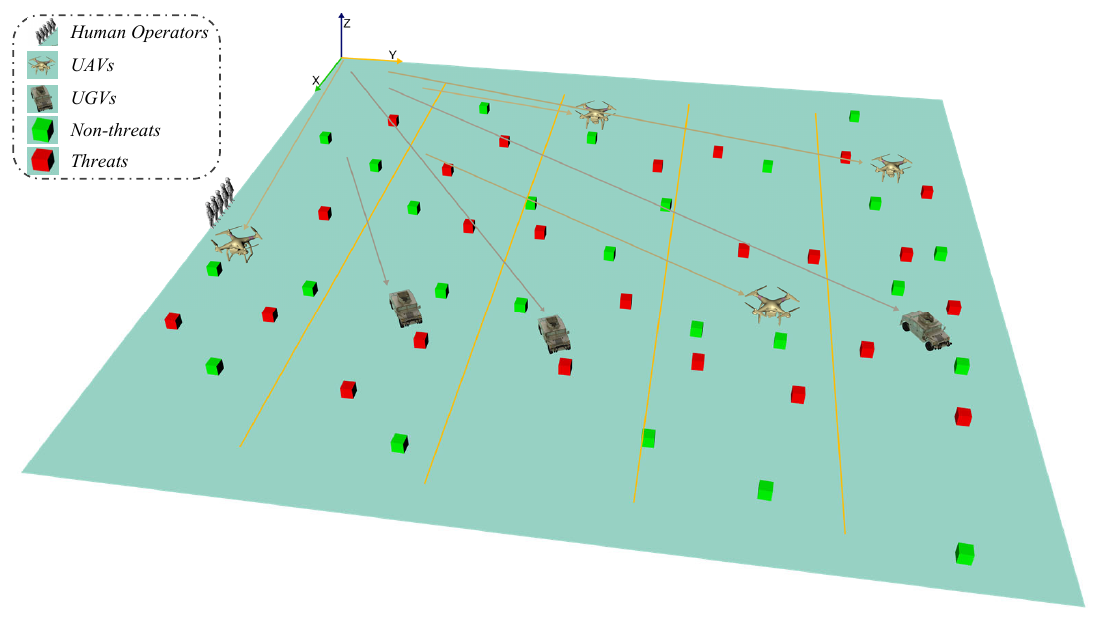}
\caption{Illustration of a simulation environment containing human operators, UAVs, UGVs, and non-threats (green cubes) and POI threats (red cubes). }
\vspace{-10pt}
\label{fig:sim}
\end{figure}

To validate our AtRL in the above task scenario, we created a simulation environment using the GAMA platform \cite{taillandier2019building}, based on the one used in \cite{humann2018modeling}. As depicted in Fig. \ref{fig:sim}, the territory to be monitored is an open space of $2~km \times 2~km$, which contains several POIs. Each POI has either a threat or a non-threat object that needs to be identified.

%, with an equal number of threats and non-threats randomly distributed across all POIs.

\subsubsection{Robot Model}
The two different robots, i.e., UAVs and UGVs, possess different specifications, such as the movement speed and quality of provided images, which are summarized in Table \ref{SIm}. However, their movement patterns and tasks are the same.

\begin{table}[h]
\centering
\caption{Specifications of UAVs and UGVs set in the simulation.}
\label{SIm}
\begin{tabular}{ccc}
\hline
\textbf{}       & UAV   & UGV \\ \hline
Speed           & 20m/s & 8m/s        \\
Image Quality   & low   & high         \\ \hline
\end{tabular}
\vspace{-5pt}
\end{table}

 At the start of each job, a robot is assigned an initial POI. The robot automatically navigates from the origin to the POI and stays there for 3 seconds to capture an image of the threat. It then publishes the image to one human operator for classification. The robot moves on to the next closest POI that has not been visited yet and repeats the image-capturing process. Once all POIs have been explored, the robot returns to the origin.

\subsubsection{Human Performance Model}
We model the human operator as an event server responsible for handling image classification tasks generated by both UAVs and UGVs. Based on the empirical findings regarding human performance in complex tasks \cite{watson2017informing}, we assume that the human server handles events serially. The performance of the binary threat identification task is assumed to be affected by the fatigue and workload of the human operator. This influence can be corrected by the individual cognitive ability of the operator. It is also affected by the inherent difficulty of the image classification task, which is subject to image quality and threat types. The difficulty level can be weighted by the individual operational skill level of the operator. The probability of correct image classification by the operator is nonlinearly influenced by these factors \cite{humann2018modeling} as:
\begin{equation}
{Pr}_c =\frac{1}{2}+ \gamma (F_f F_w) \cdot \xi (F_s)
\label{HP}
\end{equation}
\noindent where $\gamma, \xi \in (0,\frac{\sqrt{2}}{2})$  are the correction weights for individual cognitive ability and operational skill level, respectively, and $F_f, F_w\in [0,1]$, and $F_s :\in (0,1)$  denote the correction factors for fatigue, workload, and inherent task difficulty, respectively.

This formulation allows for the minimum classification success probability to be infinitely close to 0.5, which is consistent with the random guessing of a binary choice. The specific patterns and values of these correction factors are set as follows, based on \cite{humann2018modeling,tabrez2020survey}. We refer readers to these references for more details regarding the theoretical and empirical justification. The correction factors for fatigue and workload can be defined as:
\begin{equation}
F_f(\hat{t})=\left\{\begin{array}{ccl}
1 & & 0 \leq \hat{t}<1 \\
-0.12 \hat{t}+1.12 & & 1 \leq \hat{t} \leq 8
\end{array}\right.
\end{equation}
\begin{equation}
\begin{small}
F_w(u)=\left\{\begin{array}{ccl}
-2.47 u^2+2.22 u+0.5 & & 0 \leq u<0.45\\
1 && 0.45 \leq u<0.65\\
-4.08 u^2+5.31 u-0.724 && 0.65 \leq u \leq 1.0
\end{array}\right.
\end{small}
\end{equation}
\noindent where $\hat{t}$ is the working hours of a human agent, and $u$ denotes a human agent's utilization, which is the percentage of time that an operator is busy performing tasks, measured over a trailing 5-minute period.

Also, the factor that reflects the task difficulty can be defined as a sigmoid pattern \cite{pew1969speed} as:
\begin{equation}
F_s(\overline{t}) = \frac{1}{1+e^{0.05(\overline{t}-150)}}
\end{equation}
\noindent where $\overline{t}$ serves as an indicator of task difficulty, which is measured as the minimum time in seconds required to complete the task.

The value of  $\overline{t}$ is subject to the image quality and the type of threat. In our case, two levels of image quality are considered: \textit{low-quality} images captured by UAVs and \textit{high-quality} images taken by UGVs. Additionally, we consider three types of objects to be identified, presenting three levels of abstract classification difficulty. The value of $\overline{t}$ is defined in Table \ref{tab:task}.

\begin{table}[h]
\centering
\caption{Values of $\overline{t}$ under different scenarios.}
\begin{tabular}{cccc}
\hline \text { Image } & \multicolumn{3}{c}{\text { Classification Difficulty }} \\
\cline { 2 - 4 } \text { Quality } & \text { Easy } & \text { Medium } & \text { Hard } \\
\hline \text { Low } & 20 & 60 & 180 \\
\text { High } & 10 & 30 & 90 \\
\hline
\label{tab:task}
\end{tabular}
\vspace{-10pt}
\end{table}

To account for individual differences in the performance model, we introduce two correction weights: $\gamma$ and $\xi$, reflecting human cognitive abilities and operational skills, respectively. $\gamma$ is used to correct for performance decreases due to fatigue and workload, while $\xi$ is used to correct for the performance decreases due to task difficulty. The values of $\gamma$ and $\xi$ are defined as:
\begin{equation}
\gamma = sin(h_c); \xi = sin(h_s)
\end{equation}

\noindent where $h_c, h_s \in (0,\frac{\pi}{4})$ denote the individual levels of cognitive abilities and operational skills, respectively. 

We define three levels of $h_c$ and $h_s$: low $\in (0,\frac{\pi}{12})$ vs. medium $\in [\frac{\pi}{12},\frac{\pi}{6}]$ vs. high $\in (\frac{\pi}{6},\frac{\pi}{4})$. The exact value of $h_c$ or $h_s$ is randomly sampled from the range of a human's cognitive ability or operational skill level.

\subsubsection{Specifications of CMADP formulation}

In our case study, the state space, i.e., the multi-attribute context $C$, includes six sub-attributes related to tasks, robots, or humans, as summarized below:

\begin{itemize}
    \item Task attributes: location and classification difficulty of POIs
    \item Robot characteristics: speed and image quality
    \item Human factors: cognitive ability and operational skill level
\end{itemize}

The action space $A$ consists of two parts: assigning initial POIs to $j$ robots and assigning identification tasks of all POIs to $i$ humans. Note that we use K-Mean clustering to divide the POIs into $j$ areas of interest based on the location information, and regard the centroid POIs of each cluster as the candidate initial POIs to be assigned, which is consistent with the realistic military applications \cite{harriott2013modeling}. Once the initial POIs are assigned, the robots automatically navigate to the next nearest POI by taking the shortest path.

For the reward function $f_R$, we define the individual reward $r_n$ as the average performance of threat identification tasks driven by ${Pr}_c$ described in Eq. \ref{HP}. Specifically, if a POI is correctly identified, the agent will receive 10, 20, or 30 points, depending on the classification level (low to high). Conversely, if the identification is incorrect, the agent will lose the same points. Finally, the average score of all POIs obtained during one environment step is calculated as the $r_n$.

\subsection{Simulation Experiment and Results}

\subsubsection{Implementation Details}
\label{ID}
We set the number of heads in each cross-attribute attention process to 2, and each transformer's hidden unit size to 40. Moreover, the number of hidden units in the GRU was established at 64, and the GRU was set to have a single layer. For the PPO, we set the clipping ratio to 0.2, and the loss coefficients of clipped policy loss, value function loss, and entropy loss to 2, 1, and 0.1 respectively. For training, Adam optimizer is utilized and the initial learning rate is set to $2\times10^{-4}$. To boost and stabilize the training process, we use 10 behavioral actors to collect interaction experiences in parallel. The training is conducted with an NVIDIA Tesla V100 GPU. 

\subsubsection{Baselines and Ablation Model}
We compared our AtRL model with two baseline approaches: average (AV) and random (RA) allocation. The AV approach assigns initial POIs to robots in a specific order and distributes threat classification tasks equally to humans. In contrast, the RA approach assigns robots to POIs and distributes threat identification tasks randomly to humans. Additionally, to evaluate the benefits of our proposed cross-attribute attention module, we built an ablation model called RL, which removes the state representation learning part in AtRL, concatenating attribute sequences directly as the state. We trained AtRL and RL with the same PPO parameters and episodes of $1\times10^4$.

\subsubsection{Evaluation}
There are two evaluation settings: (a) 3 humans, 4 robots, 20 threats, and 20 non-threats; and (b) 5 humans, 7 robots, 25 threats, and 25 non-threats. For each setting, the attributes of humans, robots, and tasks are randomly selected to generate multiple scenarios, and 500 unseen scenarios are tested with each model. We compare the learning curves of AtRL and RL, as well as the average performance scores, i.e., the average scores obtained of threat classification tasks per run, of every model.

\begin{figure}[t]
\centering
\includegraphics[width=0.95\columnwidth]{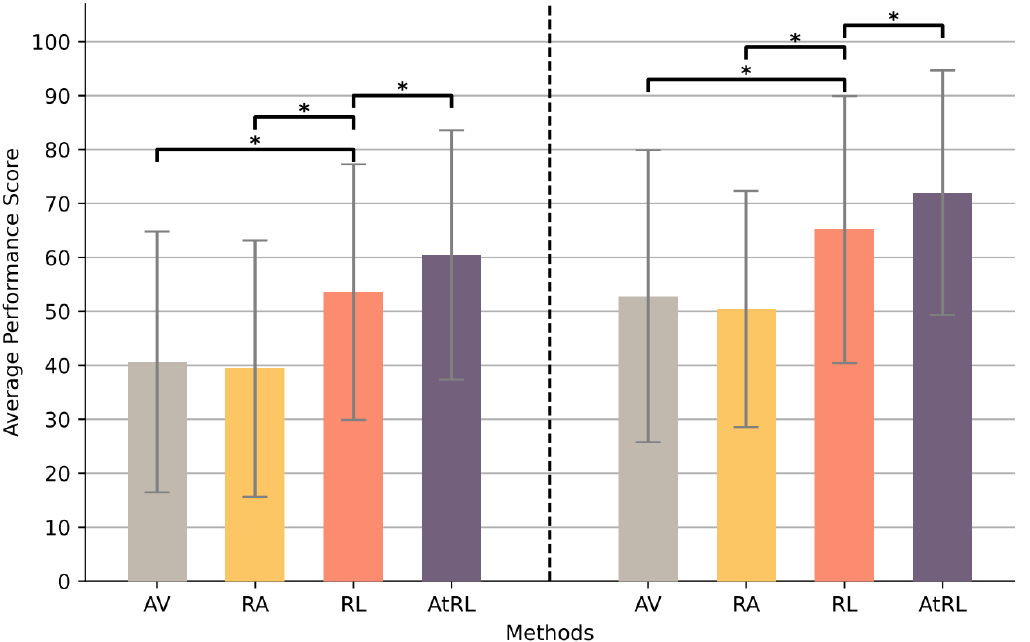}
\vspace{-5pt}
\caption{Average performance scores of each method in setting (a) 3 humans, 4 robots, 20 threats, and 20 non-threats (left), and (b) 5 humans, 7 robots, 25 threats, and 25 non-threats (right) with outcomes of two-sample t-test: $*$: $p<0.001$.}
\vspace{-10pt}
\label{fig:test}
\end{figure}

\subsubsection{Results}
\begin{figure}[t]
\centering
\includegraphics[width=0.95\columnwidth]{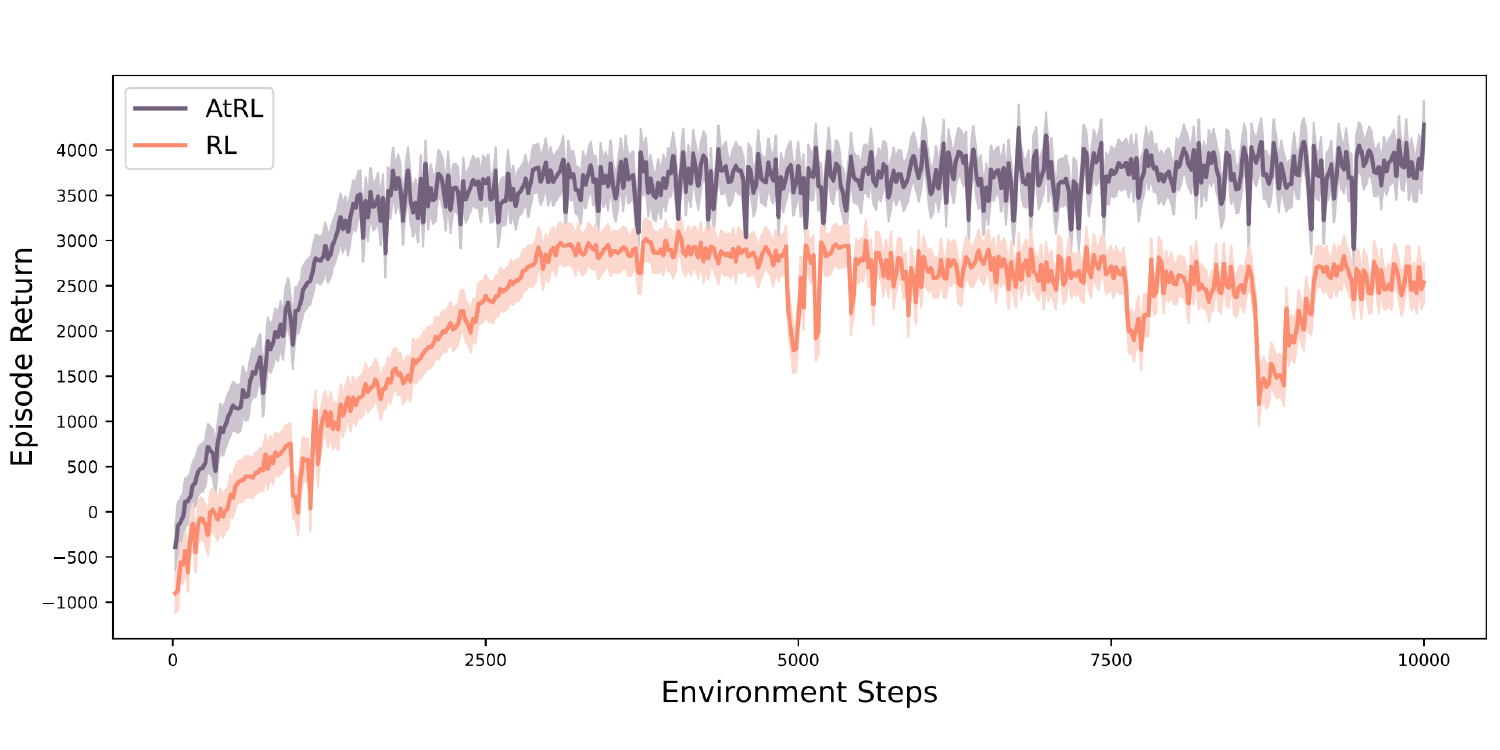}
\vspace{-8pt}
\caption{Learning curves of AtRL and RL in terms of episode return during training.}
\label{fig:curve}
\vspace{-2pt}
\end{figure}

Fig. \ref{fig:test} demonstrates the average performance scores of each run in each setting. We can observe that AtRL and RL outperform AV and RA statistically significantly ($p<0.001$). We argue the reason lies in that AtRL and RL can optimize the initial task allocation policies by continuously interacting with the multi-attribute context, while the task distribution strategies of AV and RA are fixed, neglecting the intrinsic heterogeneity among human agents, robots, and tasks. This comparison highlights the importance of the initial task allocation in a heterogeneous MH-MR team. Furthermore, AtRL outperforms RL significantly ($p<0.001$), and such performance augmentation is more obvious when the team complexity, i.e., the number of team members and tasks, is higher. This is reasonable since AtRL can learn a state representation encoded with latent dependencies hidden in the multi-attribute context. Moreover, Fig. \ref{fig:curve} shows the example learning curves of AtRL and RL during training in setting (b). We can observe that AtRL (purple) achieves more rewards with a more efficient and stable pattern compared to RL (orange). This further demonstrates the benefits of the cross-attribute state representation learning introduced in AtRL quantitatively.

\begin{figure}[t]
\centering
\includegraphics[width=0.98\columnwidth]{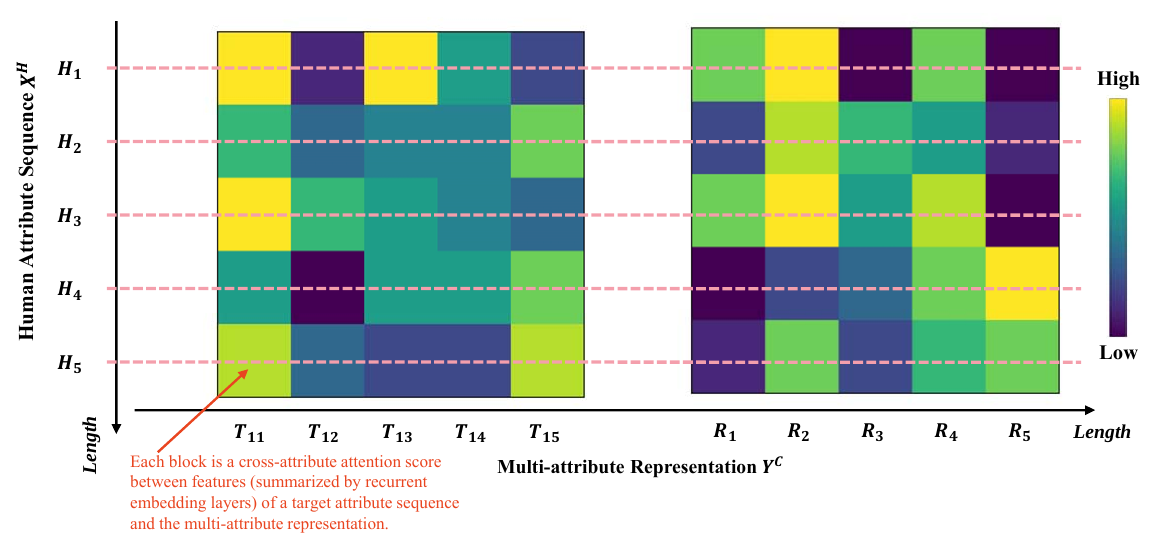}
\vspace{-3pt}
\caption{Visualization of partial example cross-attribute attention weights learned between human factor attribute sequence $X^H$ and the multi-attribute representation $Y^C$ in setting b within one batch during training. Note that the original attention weight matrix (see Fig. \ref{fig:CA}) has the dimension of $i \times c$, i.e., $5 \times 69$. }
\label{fig:visual}
\vspace{-10pt}
\end{figure}
To illustrate how our cross-attribute attention works, we visualize the attention activation. Fig. \ref{fig:visual} shows a partial example of learned cross-modal attention matrices at the final cross-attribute layer with the input of human factor sequence ${X}^H$ and the multi-attribute representation $Y^C$. We can observe that our cross-attribute has successfully learned to attend to different positions exhibiting strongly relevant information between features of humans and features of tasks and robots embedded in the multi-attribute representation in an adaptive pattern. For example, the learned cross-attention score is relatively high between $H_3$ which presents a human operator with a low level of cognitive ability, and $T_{11}$ which denotes an image classification task at a far location as well as $R_2$ which is a UGV with a low speed. This is reasonable as cognitive ability decides the human resistance to fatigue that is sensitive to work time, which in our case is influenced by the navigation speed of robots and the location of the POI.

\section{Conclusion}
In this work, we investigated the initial workload allocation problem in a multi-human multi-robot team taking the inherent heterogeneity of humans, robots, and tasks into account. We provided a general problem formulation of a contextual multi-attribute decision-making process and proposed a novel attention-based deep reinforcement learning approach. To capture the complex dependencies across multiple attributes and sub-attributes in a multi-attribute context, we introduced a cross-attribute attention module for state representation learning. A comprehensive case study under the scenario of a threat surveillance task was conducted to demonstrate the benefits of our approach and the proposed module inside. In future work, we intend to test our method in more complex scenarios, such as cluttered environments. We also plan to conduct a user study to refine our human model and validate the effectiveness of the AtRL approach in real-world applications. Additionally, our current work is aiming to enhance our model's adaptability within more large-scale MH-MR teams.

\section*{ACKNOWLEDGMENT}
This paper is based on research supported by the National Science Foundation (NSF) under Grant No. IIS-1846221. Any opinions, findings, and conclusions or recommendations expressed in this material are those of the authors and do not necessarily reflect the views of the National Science Foundation.

%\vspace{-10pt}
\typeout{}
\bibliography{main}
\bibliographystyle{IEEEtran}
\end{document}